# A Comparative Study on Different Types of Approaches to Bengali document Categorization

**Md. Saiful Islam[1,*], Fazla Elahi Md Jubayer** [1] **and Syed Ikhtiar Ahmed** [1]

[1] Department of Computer Science and Engineering, Shahjalal University of Science & Technology, Sylhet, Bangladesh..

**Abstract**: Document categorization is a technique where the category of a document is determined. In this paper three well-known supervised learning techniques which are Support Vector Machine(SVM), Naïve Bayes(NB) and Stochastic Gradient Descent(SGD) compared for Bengali document categorization. Besides classifier, classification also depends on how feature is selected from dataset. For analyzing those classifier performances on predicting a document against twelve categories several feature selection techniques are also applied in this article namely Chi square distribution, normalized TFIDF (term frequency-inverse document frequency) with word analyzer. So, we attempt to explore the efficiency of those three-classification algorithms by using two different feature selection techniques in this article.

## 1. INTRODUCTION

Content based classification is emerging nowadays due to continuous growth of electronic data. It is the process of grouping documents into different classes or categories. So document categorization plays an important role in natural language processing, computer science and information science. In digital library system, search engine, and document management system this automatic categorization can be used. Spam filtering (Sahami et al., 1998 ), online news filtering (Chan et al., 2008 ), social media analytics (Melville et al., 2009), survey data grouping etc. are some applications of document categorization.

Extensive research has already been done in this field for various languages. There are several powerful techniques offered by natural language processing. Three ways through which a document can be classified are unsupervised, supervised and semi-supervised techniques. In this paper we use supervised learning. Supervised learning algorithm is a technique which analyzes labeled training data sets and produces a function by which it can map new examples for prediction. The main focus of this article is categorize Bangla documents. Bangla is one of the most popular languages throughout the world. According to the consensus of the total number of native speakers, it is the 7th most spoken language (Islam, 2009). So it is very much needed to organize and categorize Bengali documents automatically so that users can easily find required and related information. For this purpose, this paper describes how to automatically categorize Bengali documents using the supervised learning technique. The result achieved for twelve categories by using the formula described later in this paper, is quite promising as it is better than the existing research on Bangla document classification.

---

* Corresponding author: * saiful-cse@sust.edu

## 2. RELATED WORKS

From literacy review we see that many research works have been done in text or document categorization for English languages. Lots of well-established supervised techniques are used most frequently such as Denoeux (1995) used KNearest Neighbor(KNN), Chen et al. (2009) used Naïve Bays(NB), Brown et al.(1992) used N-grams,Quinlan (1986) used Decision Tree(DT), Sebastiani (2002) used Neural Network(NNet), Cortes and Vapnik (1995) used SVM. There are fair amounts of comparative.

study also done in English DC. Patil and Pawar (2012) used Naive Bayes algorithm for classify the content in a web sites. They got average 80% accuracy for ten categories. Bijalwan et al. (2014) used KNN, NB and Term-gram for this task. In their experiment they showed that the accuracy of KNN is better choice than NB and Term-gram. Besides this, Tam et al. (2002) also showed that KNN is performed better than NNet and NB for English document. Y et al. (2012) did an comparative study on DT, NB, KNN, Rocchio's Algorithm, Back propagation Network, SVM. In their comparisons they showed that SVM is performed far better than all other approaches they used for 20Newsgroups dataset. Also Zhijie et al. (2010) compare SVM against KNN and NB classifier and their statistics proved that SVM is better than KNN and NB. Joachims (1998) was the first who propose the use of a linear SVM with TFIDF term feature for DC. Zakzouk and Mathkour (2011) used SVM for classify cricket sports news.

Besides English document there also done good amount of research on other languages too. KOURDI et al. (2004) used SVM , Mesleh (2008) used NB for automatic text classification in Arabic languages. For Tamil (South Indian language) languages Rajan et al. (2009) used Naïve Bayes and Gupta (2012) used Neural networks for this task.

In addition, few works have been done on document categorization for Bangla languages. Mandal and Sen(2014) compared four supervised learning techniques for labeled web documents into five categories.

## 3. FEATURE SELECTION ALGORITHM:

In this section, we briefly describe TF-IDF and Chi-Square distribution methods used in this study.

**TF-IDF:**

Textual representation is converted to vector space model consisting of term frequency. The fundamental issue with the term-frequency method is that it scales up frequent terms as a result importance of rare terms is ignored, though low frequency terms are significant for distinguish between classes. Where TFIDF approaches scale down the frequent terms while scaling up the uncommon or rare terms. If a term occurs in 10 documents among total 15 documents then it is not as important as which is occur less than 10 document or occur only one document. That's why TFIDF uses the logarithm scale to do that tricks. Next, IDF computed using the following formula:

$$\ln\left(\frac{N+1}{DF+1}\right)+1$$

**CHI-SQUARE DISTRIBUTION:**

Chi-square distribution is a simple statistical approach. In paper[25] authors used this method to determine important words in a document. Mathematically chi-square distribution can be written as:

$$\chi^2 = \sum_{i=0}^{n} \frac{\{o_i - \varepsilon_i\}^2}{\varepsilon_i}$$

Where $X^2$ represents chi-square value, $o_i$ represents observed frequency, $\varepsilon_i$ represents expected frequency and degree is freedom is 1.
For calculating chi value of each term in a document the equation can be modified as:

$$X^2(w) = \sum_{g \epsilon G} \frac{\{freq(w,g) - P_g n_w\}^2}{P_g n_w}$$

Where, $P_g$ = TF of a word/ total number of word in the document.
  $n_w$  = TF of a word w in a sentence where w appears.
  $P_g n_w$ = Frequency of word co-occurrence.
So, if a word appears in a long length sentence in a document it threated as an important term as they those words likely to co-occur with many other terms.

## 4. CLASSIFICATION ALGORITHM:

### NAÏVE BAYES:

Naïve Bayes algorithm is one kind of probabilistic approach, based on applying Bayes theorem. Contingent upon the exact way of the likelihood show, this classifier can be prepared productively in a supervised learning methods.

### STOCHASTIC GRADIENT DESCENT:

Stochastic Gradient Descent (SGD) is a straightforward yet extremely productive way to deal with discriminative learning of linear classifiers under curved misfortune capacities, for example, (linear) Support Vector Machines and Logistic Regression [26]. Despite the fact that SGD has been around in the machine learning group for quite a while, it has gotten a lot of consideration only as of late with regards to expansive scale learning. It is effectively applied to large-scale and sparse machine learning issues frequently experienced in DC. For sparse data, it easily scale to problems with more than 10^5 training samples and more than 10^5 features.

### SUPPORT VECTOR MACHINE:

Another supervised learning algorithm is Support Vector Machine(SVM) which is widely and effectively used for DC. While training a classifier it needs to manage a lot of features. For our case it was 216576 features from 28717 train documents (table II) which was calculated after preprocessing and feature selection. Since SVMs use overfitting protection, which does not necessarily depend on the number of features, they have the potential to handle these large feature spaces. One of the important properties of text categorization is that most of the DC problems are linearly separable. SVMs algorithm can able to find such linear as well as polynomial, RBF separator. Basically SVM act as binary classifier. If n is the number of total features, then SVM plot each data item as a point in n-dimensional space where the value of each features represent the value of a particular coordinate. Then for separate two classes performing this classification algorithm find a hyperplane. More specifically, a SVM create a hyperplane or a set of hyperplane in a n-dimensional space, which can be used to classify. Theoretically, if x is the training data set and w is n weighted vector in $R^n$ then the learning function of SVM is f(x) = sign(wx + b).

## 5. EXPERIMENTAL SETUP AND RESULTS:

### DATASET USED IN THIS EXPERIMENT:

All the documents that are used in these experiments are collected from a Bengali document corpus [27]. All documents are collected from various Bangladeshi newspapers such as http://prothom-alo.com, http://bdnews24.com, http://dailykalerkantha.com etc. and are labeled with their corresponding category name. From this corpus, we collect almost same number of documents for training and testing for each category which are represented in table I. There are twelve categories.

Table 1. Total 31908 samples

| Category | Train/Test |
|---|---|
| Accident(A) | 2393/266 |
| Art(Ar) | 2370/289 |
| Crime(C) | 2402/257 |
| Economics(E) | 2386/273 |
| Education(Ed) | 2401/258 |
| Entertainment(En) | 2426/233 |
| Environment(Env) | 2371/288 |
| International(I) | 2402/257 |
| Opinion(O) | 2403/256 |
| Politics(P) | 2395/264 |
| Science and Technology(S) | 2371/288 |
| Sports(Sp) | 2397/262 |
| **Total** | **28717/3191** |

**PREPROCESS DATASET:**

Before using the documents, training the classifier model preprocessing data set is a crucial step. It is needed because it removes recurring words or symbols from each document which are common for all documents and plays an important role in dimension reduction of feature space and increase in the performance. So by doing this, it will find those important words only which are relevant for the information of data set, from which classifier can determine the label of document. The steps are (i)Tokenization (ii) Remove frequent symbol (iii) Stemming and (iv) Remove all pronouns and conjunction.

**APPLY FEATURE SELECTION AND TRAIN MODEL:**

After preprocess done, two different feature engineering techniques as described above were applied separately with SVM, SGD, and NB classifier and compare performance with each other.

For chi-square distribution system, we sorted all features in a document in ascending order after this top 30% feature selected. Total 91503 features selected from total 28717 training documents using preprocess+chi-square method. For TFIDF we used unigram as word analyzer and normalized it weighting with length normalization. Total 216576 features selected from the same size train documents as previously used while using preprocess+TFIDF which is almost 2.3 times higher number of features then we found from preprocess+chi-square method.

While using NB classifier, alpha value fixed to 0.01 and SGD classifier tuned with hinge as loss function, l2 regularization as penalty, 0.0001 as alpha value and 50 as number of iteration. For implementing SVM, LIBSVM was used which is a library for SVM. It supports multiclass classification. With the help of this library we used C-SVC method with linear kernel for classifier.

Using table I training data set and above configuration we performed CHI-SQUARE+NB, CHI-SQUARE+SGD, CHI-SQUARE+SVM, TFIDF+NB, TFIDF+SGD, TFIDF+SVM.

precision, recall. F-measure(macro average) and accuracy are calculated for those methodologies. Precision represents how many labels were correctly predicted (for a given class A) from all predicted label. Recall represents how many labels were correctly predicted from all instances that should have a label A. F-measure represents the weighted average of the precision and recall.

5 | Md. Saiful Islam et. al., I C E R I E 2 0 1 7Table 2. Performance of different approaches

|  | Train Time(sec) | Precision (%) | Recall (%) | F1-Measure (%) |
|---|---|---|---|---|
| CHI-SQUARE+SGD | 6.21 | 83.90 | 83.81 | 83.56 |
| TFIDF+SGD | 14.44 | 92.06 | 92.11 | 92.07 |
| CHI-SQUARE+NB | 0.086 | 83.67 | 83.34 | 83.36 |
| TFIDF+NB | 0.2370 | 89.59 | 89.45 | 89.48 |
| CHI-SQUARE +SVM | 151.68 | 84.49 | 84.41 | 84.34 |
| TFIDF+SVM | 469.09 | 92.56 | 92.58 | 92.57 |

## 6. CONCLUSION:

After calculating the performance of different approaches it is found that SVM classifier obtained the highest F1-score(92.56%) while normalized TFIDF used as feature selection and CHI-SQUARE+NB was the lowest with F1-score 83.36% . It is also observed that all the classifier performed better with TFIDF feature selection technique than chi-square distribution method. Below chart shows the comparisons between them:

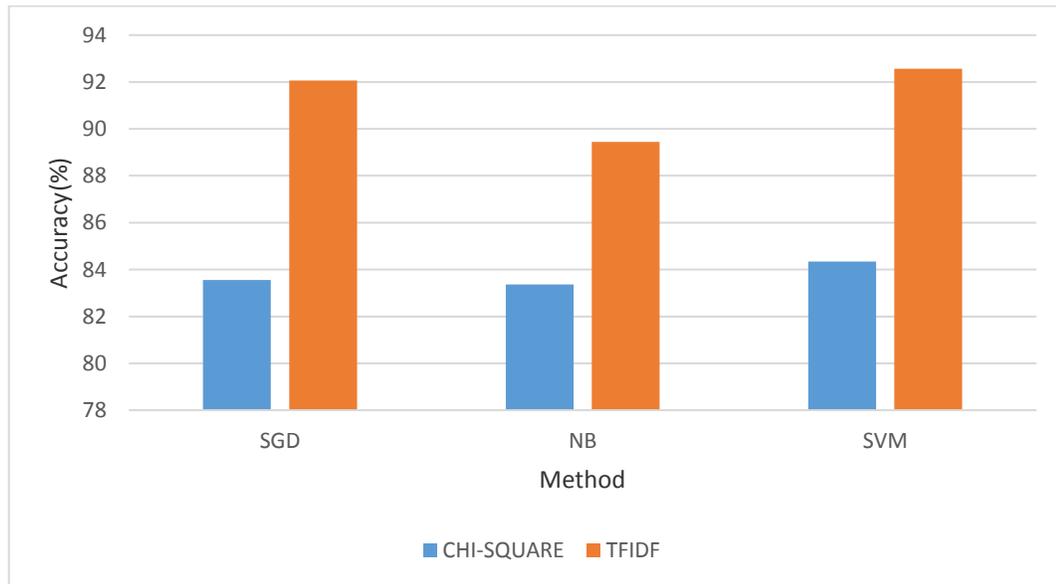

**Figure I.** Comparisons of different approaches.

## 8. REFERENCES

M. Sahami, S. Dumais, D. Heckerman, and E. Horvitz. A Bayesian approach to filtering junk email., AAAI Workshop on Learning for Text Categorization, July 1998, *Madison, Wisconsin. AAAI Technical Report* WS-98- 05.

C. Chan, A. Sun, and E. Lim, "Automated Online News Classification with Personalization," *in 4th International Conference of Asian Digital Library*, 2001.

P. Melville, V. Sindhwani, and R. Lawrence, "Social media analytics: Channeling the power of the blogosphere for marketing insight.," *in Workshop on Information in Networks*, 2009.

M. Islam, "Research on Bangla language processing in Bangladesh: progress and challenges," *8th ILDC, Dhaka, Bangladesh* (June 2009), 2009.

T. Denoeux, "A k-nearest neighbor classification rule based on Dempster-Shafer theory," *Systems, Man and Cybernetics, IEEE Transactions* on, vol. 25, pp. 804-813, 1995.

J. Chen, H. Huang, S. Tian, and Y. Qu, "Feature selection for text classification with Naïve Bayes,"Expert Systems with Applications, vol. 36, pp. 5432-5435, 2009.

P. F. Brown, P. V. Desouza, R. L. Mercer, V. J. D. Pietra, and J. C. Lai, "Class-based n-gram models of natural language," *Computational linguistics*, vol. 18, pp. 467-479, 1992.

J. R. Quinlan, "Induction of decision trees," *Machine learning*, vol. 1, pp. 81-106, 1986.

F. Sebastiani, "Machine learning in automated text categorization," *ACM computing surveys (CSUR)*, vol. 34, pp. 1-47, 2002.

C. Cortes and V. Vapnik, "Support-vector networks," *Machine learning*, vol. 20, pp. 273-297, 1995.

A. S. Patil and B. Pawar, "Automated classification of web sites using Naive Bayesian algorithm," *in Proceedings of the International MultiConference of Engineers and Computer Scientists*, 2012, pp. 14-16.

V. Bijalwan, V. Kumar, P. Kumari, and J. Pascual, "KNN based Machine Learning Approach for Text and Document Mining," *International Journal of Database Theory and Application*, vol. 7, pp. 61-70, 2014.

Tam, Santoso A and Setiono R., "A comparative study of centroid-based, neighborhood-based and statistical approaches for effective document categorization*", ICPR '02 Proceedings of the 16 th International Conference on Pattern Recognition (ICPR'02)*, vol.4, no. 4, 2002, pp.235–238.

Pratiksha Y. Pawar and S. H. Gawande, "A Comparative Study on Different Types of Approaches to Text Categorization" *International Journal of Machine Learning and Computing*, Vol. 2, No. 4, August 2012

L. Zhijie, L. Xueqiang, L. Kun, and S. Shuicai, "Study on SVM Compared with the other Text Classification Methods," *in Education Technology and Computer Science (ETCS), 2010 Second International Workshop* on, 2010, pp. 219-222.

Thorsten Joachims. 1998. Text categorization with Support Vector Machines: Learning with many relevant features. *In Proceedings of the 10th European Conference on Machine Learning*, ECML'98, pages 137–142.

T. Zakzouk and H. Mathkour, "Text Classifiers for Cricket Sports News," *in proceedings of International Conference on Telecommunications Technology and Applications ICTTA*, 2011, pp. 196-201.

Mohamed EL KOURDI, Amine BENSAID, Tajje-eddine RACHIDI, "Automatic Arabic Document Categorization Based on the Naïve Bayes Algorithm," *in proceedings of the Workshop on Computational Approaches to Arabic Script-based Languages*, 2004, pp. 51-58.

A. Moh'd Mesleh, "Support vector machines based Arabic language text classification system: feature selection comparative study," *in Advances in Computer and Information Sciences and Engineering*, ed: Springer, 2008, pp. 11-16.

K. Rajan, V. Ramalingam, M. Ganesan, S. Palanivel, and B. Palaniappan, "Automatic classification of Tamil documents using vector space model and artificial neural network," *Expert Systems with Applications*, vol. 36, pp. 10914-10918, 2009.

N. a. V. Gupta, "Algorithm for Punjabi Text Classification," *International Journal of Computer Applications*, vol. 37, pp. 30-35, 2012.

Ashis Kumar Mandal, Rikta Sen, "Supervised Learning Methods For Bangla Web Document Categorization," *International Journal of Artificial Intelligence & Applications (IJAIA)*, Vol. 5, No. 5, September 2014

Y. Matsuo, and M. Ishizuka, "Keyword Extraction from a Single Document using Word Co-occurrence Statistical Information", *International journal on Artificial Intelligence Tools*, vol.13, no.1, pp.157-169, 2004.

T. Zhang, "Solving large scale linear prediction problems using stochastic gradient descent algorithms," *in Proceedings of the Twenty-first International Conference on Machine Learning, ser. ICML '04. New York, NY, USA: ACM*, 2004, pp. 116–. [Online]. Available: http://doi.acm.org/10.1145/1015330.1015332.

http://scdnlab.com. Bangla Corpus[Online]. Available: http://scdnlab.com/corpus/ [Accessed: Retrieved Oct 13, 2016].